# Measuring similarity between geo-tagged videos using largest common view

Wei Ding, KwangSoo Yang and Kwang Woo Nam✉

This paper presents a novel problem for discovering the similar trajectories based on the field of view (FoV) of the video data. The problem is important for many societal applications such as grouping moving objects, classifying geo-images, and identifying the interesting trajectory patterns. Prior work consider only either spatial locations or spatial relationship between two line-segments. However, these approaches show a limitation to find the similar moving objects with common views. In this paper, we propose new algorithm that can group both spatial locations and points of view to identify similar trajectories. We also propose novel methods that reduce the computational cost for the proposed work. Experimental results using real-world datasets demonstrates that the proposed approach outperforms prior work and reduces the computational cost.

*Introduction:* Recent advances in mobile technologies such as GPS, accelerometers, gyro sensors, and dash cams have significantly increased the quantity of GeoVideo generated by mobile users. GeoVideo contains a variety of spatial properties, such as the latitude and longitude at which the video was taken, the direction of the view in which the video was taken, and the angle of the camera. These spatial properties enable us to identify new interesting trajectory patterns that can utilize geometric and topological locations, regions of interests, and direction of user's views. Consider the example in Fig. 1. Three people follow the same direction toward a tourist attraction and capture videos. Which trajectories are similar? Prior work, such as Hausdorff or Longest Common Subsequence (LCSS) approaches [1,2], consider only either spatial locations or spatial relationship between two line-segments and show that GeoVideo2 and GeoVideo3 are the most similar. Is it even semantically true? Even though GeoVideo2 and GeoVideo3 have a similar trajectory, the field of views (FoVs) of the two cameras are different, which results in semantically incorrect output. Therefore, in terms of semantic similarity, GeoVideo1 and Geovideo3 should be the most similar due to similar FoVs. To satisfy this requirement, we propose novel approach, namely Largest Common View Subsequence (LCVS), which measures similarity between trajectories based on the Field of View (FoV).

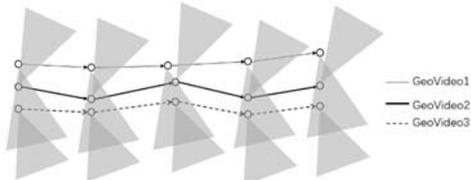

**Fig. 1** *An example of the geo-tagged videos with similar trajectories, but different view directions. Geovideo 1 and 2 are similar trajectory distance but in terms of view of the video, geovideo 1 and 3 are more similar.*

The proposed approach is important for many societal applications such as grouping moving objects, classifying geo-images, and identifying the interesting trajectory patterns. For example, suppose you have a video of many parents and children groups moving hand in hand at a mall. Prior work, such as Hausdorff and LCSS, may produce similar trajectories, but show a limitation to distinguish groups based on common interests. Since our proposed approach considers FoV in the similarity measurement, it can clearly identify similar group of trajectories based on common interests (e.g., viewpoints).

*Background:* Prior work on identifying the similar trajectories can be categorized into two groups: 1) distanced-based approach and 2) common subsequence-based approach. Distance based approaches use distance metrics (e.g., Euclidean or Hausdorff distance) to measure the similarity between two trajectories [1]. However, such approaches do not guarantee the accuracy due to outliers or different speeds in trajectories. Common subsequence-based approaches use the longest common subsequence to identify the similar trajectories [2]. However, these approaches have not been designed to account for the field of the view or the angle of the camera. By contrast, this paper proposes a novel approach for identify semantically interesting group of trajectories that considers both spatial locations and the direction of viewpoints.

Our proposed approach utilizes the concept of Longest Common SubSequence problem (LCSS) [3]. LCSS uses the distance threshold and the time threshold between two spatial points to identify a common subsequence [3]. The distance threshold is used to find the common locations. If the distance between two spatial locations is within the distance threshold, then two spatial locations are the same. The main issue of the LCSS method is that it may produce semantically incorrect results because spatial locations within the distance threshold may have different or opposite viewpoints. To remedy this issue, our approach uses the weighted distance measurement based on the common views.

*Largest Common View Subsequence (LCVS):* GeoVideo consists of a series of video frames, associated with spatial properties. Fig. 2 shows an example of GeoVideo. Let $i$ be the timestamp, let $p_i$ be the location of the camera, let $r_i$ be the maximum visible distance from the camera, let $\theta_i$ be the angle from the north to the direction of the camera, and let $\delta_i$ be the maximum horizontal angle of the camera lens. In this example, the video frame at timestamp $i$ is associated with a set of spatial properties (i.e., $p_i$, $r_i$, $\theta_i$, and $\delta_i$). Let $fov_i$ be a set of spatial properties (i.e., $fov_i=(p_i, r_i, \theta_i, \delta_i)$) and let $View(fov_i)$ be the region of $fov_i$. Given two FoV regions (e.g., $View(fov_i)$ and $View(fov_j)$), the weight of the common view (CVW) between $View(fov_i)$ and $View(fov_j)$ is defined as follows:

$$CVW(fov_i, fov_j) = \frac{|View(fov_i) \cap View(fov_j)|}{|View(fov_i) \cup View(fov_j)|}, \quad (1)$$

where $|View(fov_i) \cup View(fov_j)|$ is the union of two FoV regions and $|View(fov_i) \cap View(fov_j)|$ is the intersection of two FoV regions.

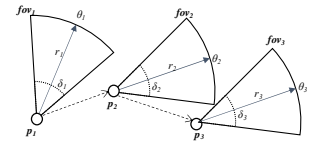

**Fig. 2** *Example of FoV model in GeoVideo*

Assume that GeoVideo $A$ consists of $m$ number of FoVs (i.e., $A = \{fov_1, fov_2 ..., fov_m\}$) and GeoVideo $B$ consists of $n$ number of FoVs (i.e., $B = \{fov_1, fov_2 ..., fov_n\}$). Let $Head(A)$ be a sequence of FoVs of $A$ from $1$ to $m-1$. (i.e., $Head(A) = \{fov_1, ..., fov_{m-1}\}$) and let $\sigma$ be the minimum time threshold. The maximum common view-based similarity distance is defined as

$$LCVS_\sigma(A,B) = \begin{cases} 0 & \text{if } A \text{ or } B \text{ is empty}, \\ CVW(A.fov_m, B.fov_n) + LCVS_\sigma(Head(A), Head(B)) \\ \quad \text{if } CVW(A.fov_m, B.fov_n) > 0 \\ \quad \text{and } |n-m| \le \sigma \\ \max\{LCVS_\sigma(Head(A), B), LCVS_\sigma(A, Head(B))\} \\ \quad \text{otherwise} \end{cases} \quad (2)$$

*LCVS similarity and distance function:* Given a minimum time threshold $\sigma$, the LCVS similarity between A and B is defined as

$$Similarity(A,B,\sigma) = \frac{LCVS_\sigma(A,B)}{min(m,n)} \quad (3)$$

The LCVS distance between A and B is defined as

$$Distance(A,B,\sigma) = 1 - Similarity(A,B,\sigma) \quad (4)$$

The LCVS distance is metric because it satisfies the following three properties:
1) $Distance(A,B,\sigma) \ge 0$ for all $A, B \ne \emptyset$ (non-negativity)
2) $Distance(A,B,\sigma) = Distance(B,A,\sigma)$ for all $A, B \ne \emptyset$ (symmetry)
3) $Distance(A,B,\sigma) + Distance(B,C,\sigma) \ge Distance(A,C,\sigma)$ for all $A, B, C \ne \emptyset$ (triangle inequality)

*Algebraic Cost Model of LCVS*: LCVS enumerate all sub-sequences of GeoVideo $A$, which takes $O(2^m)$. For each sub-sequence, LCVS scans $n$ number of FoVs in GeoVideo $B$ to compute $LCVS_\sigma(A,B)$. This takes $O(n)$. Therefore, the cost model of LCVS is $O(n \cdot 2^m)$.

*Implementation and Experimental Result*: The main performance bottleneck of the proposed approach is to compute the FoV region, the union of two FoV regions, and the intersection of two FoV regions. To improve the efficiency of LCVS, we employ three novel methods to reduce the computational cost of computing the weight of the common view (CVW). Our first method uses Minimum Bounding Segment (MBS) to approximate the FoV region (see Fig.3*a*). This method partitions FoV using the same-sized triangles and sums up the area of these triangles to estimate the FoV region, which takes a linear time in terms of the number of triangles. Our second method uses Minimum Bounding Triangle (MBT) to speed up the computation of the FoV region (see Fig. 3*b*). Our third method uses the Minimum Bounding Rectangle (MBR) to roughly estimate the FoV region (see Fig. 3*c*), which can significantly reduce the computation cost for CVW in a special case (e.g., car dash cams).

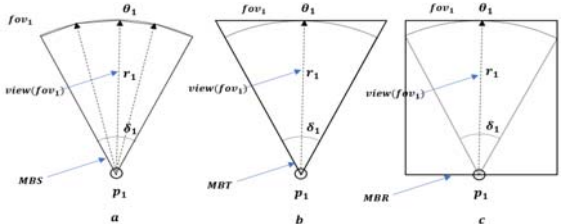

**Fig. 3** *Simplification of a FoV region using MBS, MBT, MBR*
*a* MBS, FoV is approximated by a set of segments (or triangles)
*b* MBT, FoV is approximated by a triangle
*c* MBR, FoV is approiximated by an rectangle

We conducted experiments to evaluate performance of the LCVS algorithm. We tested four different approaches: 1) Original LCSS, 2) LCVS with MBS, 3) LCVS with MBT, and 4) LCVS with MBR. The overall goal was to show the performance improvements of our new three methods to identify similar FoV trajectory patterns based on geo-tagged video datasets. We wanted to answer two questions: (1) What is the effect of the number of FoVs? (2) What is the effect of viewable distances? Performance measurements were the accuracy of outputs and execution time.

We used geo-tagged 4,000 real-world driving videos in New York City, taken from BDD100K [4]. We generated two different datasets: 1) straight-ahead direction FoV and 2) random direction FoV. In straight-ahead direction FoV dataset, the direction of camera is aligned with the direction of moving object, which can be obtained from dashcam recordings. In random direction FoV dataset, the direction of camera is randomly varying without an alignment with the direction of moving object, which can be obtained from mobile device camera recordings (e.g., smart phone). In our experimental setup, we fixed the minimum time threshold σ to 1 and the angle of a segment (i.e., the acute angle of a triangle) in MBS to 5.

The first experiment evaluated the effect of the number of FoVs. The number of FoVs was varied from 1,000 to 4,000. Fig. 4*a* and 4*b* give the accuracy of outputs. As can be seen, our approaches outperform the original LCSS. The performance gap increases as the number of FoVs increases. This is because the original LCSS considers only point distance to measure the similarity between two trajectories. Fig. 4*b* shows that LCVS with MBS and LCVS with MBT outperform LCVS with MBR. This is because MBR can roughly estimate the FoV region without considering the accuracy of outputs.

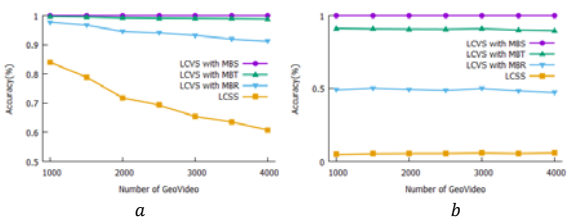

**Fig. 4** *Effect of the number of FoVs*
*a* Strait-ahead direction FoV dataset
*b* Randon direction FoV dataset

The second experiment evaluated the effect of viewable distances. We varied the distance interval from 10m to 60m. Fig. 5*a* and 5*b* give the accuracy of outputs. As can be observed, both LCVS with MBS and LCVS with MBT outperform other approaches. Fig. 5*b* shows that LCVS with MBR performs poorly because the error field is increased with viewable distance in random direction dataset.

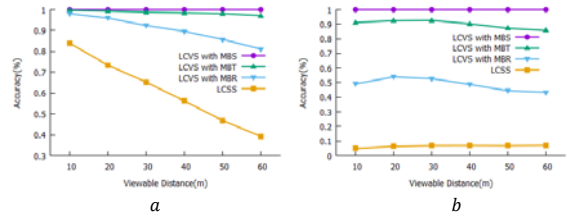

**Fig. 5** *Effect of viewable distances*
*a* Strait-ahead direction FoV dataset
*b* Randon direction FoV dataset

The third experiment evaluated the runtime of the LCVS algorithm. We varied the number of FoVs from 1,000 to 4,000 (see Fig. 6*a* and 6*b*) The results show that the original LCSS performs slightly better than LCVS with MBT and LCVS with MBR. This is because our approaches compute CVW to measure the similarity between two trajectories. However, given that the accuracy of the original LCSS is much lower than our approaches, choosing LCVS to identify similar FoV trajectory patterns is better choice. We can see that LCVS with MBS is slower than other approaches because MBS takes a linear time to compute CVW.

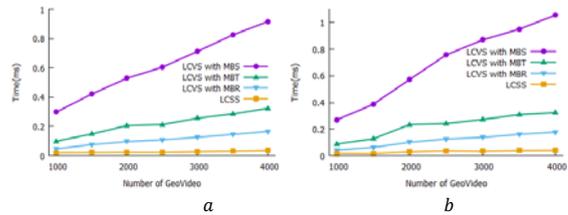

**Fig. 6** *Effect of the number of FoV*
*a* Strait-ahead direction FoV dataset
*b* Randon direction FoV dataset

*Conclusion:* We described a novel approach to group both spatial locations and points of view to identify similar trajectories. We proposed three new methods to improve the performance of the proposed work. Experimental results using real world dataset demonstrated that our proposed approach outperforms related work and reduces the computational cost.

*Acknowledgments:* This work has supported by the National Research Foundation of Korea grant (No.2018R1A2B6007982), and National Land Space Information Research Program grant(14NSIP-B080144-01) of Korean government.

Wei Ding and Kwang Woo Nam (*School of Computer, Information, and Communications Engineering, Kunsan National University, Kunsan, Republic of Korea*); KwangSoo Yang (*Department of Computer Science, Florida Atlantic University, Boca Raton, Florida, United States*)
E-mail: kwnam@kunsan.ac.kr

### References

1. Gastaldo, P., Zunino, R.: 'Hausdorff distance for robust and adaptive template selection in visual target detection', *Electronics Letters*, 2002, 38, (25), pp. 1651-1653.
2. Vlachos, M., Gunopulos, D., Kollios, G.: 'Discovering similar multidimensional trajectories', *International Conf. of Data Engineering*, San Jose, February 2002, pp. 673-684.
3. Atev, S., Miller, G., Papanikolopoulos, N. P.: 'Clustering of vehicle trajectories', IEEE Trans. *Intelligent Transportation Systems*, 2010, 11, (3), pp. 647-657.
4. Yu, F., Xian, W., Chen, Y., Liu, F., Liao, M.: 'BDD100K: A diverse driving video database with scalable annotation tooling', *arXiv preprint*, 2018, arXiv:1805.04687.